\documentclass{article}
\usepackage{style}
\usepackage{times}

\usepackage{amsmath,amsfonts,bm}

\def\eqref#1{equation~\ref{#1}}

\def\1{\bm{1}}

\DeclareMathAlphabet{\mathsfit}{\encodingdefault}{\sfdefault}{m}{sl}
\SetMathAlphabet{\mathsfit}{bold}{\encodingdefault}{\sfdefault}{bx}{n}

\usepackage{amssymb}
\usepackage{amsmath}
\usepackage{amsthm}
\usepackage{bbm}
\newtheorem{theorem}{Theorem}
\newtheorem{corollary}{Corollary}[theorem]

\theoremstyle{definition}
\newtheorem{definition}{Definition}
\newtheorem{example}{Example}

\usepackage{tikz}
\usetikzlibrary{arrows,positioning,shapes.misc}
\usepackage{pgfplots} 
\pgfplotsset{compat=1.8}
\usepackage{nicefrac}
\usepackage{algorithm}
\usepackage{algorithmic}
\usepackage{xcolor}
\usepackage{comment}
\usepackage{booktabs}
\usepackage{authblk}

\usepackage{url,hyperref,microtype,subcaption}

\author[1]{Thomas Kehrenberg}
\author[1,2]{Zexun Chen}
\author[1]{Novi Quadrianto}
\affil[1]{Predictive Analytics Lab (PAL), University of Sussex, Brighton, UK}
\affil[2]{BioComplex Laboratory, University of Exeter, Exeter, UK}

\title{Tuning Fairness by Balancing Target Labels} 

\newcommand{\PP}{\mathbb{P}} 

\iclrfinalcopy
\begin{document}



\maketitle

\begin{abstract}

%
The issue of fairness in machine learning models has recently attracted a lot of attention as ensuring it will ensure continued confidence of the general public in the deployment of machine learning systems.
We focus on mitigating the harm incurred by a biased machine learning system
that offers better outputs (e.g.\ loans, job interviews) for certain groups than for others.
We show that bias in the output can naturally be controlled in probabilistic models
by introducing a latent target output. 
This formulation has several advantages:
first, it is a unified framework for several notions of group fairness such as Demographic Parity and Equality of Opportunity;
second, it is expressed as a marginalisation instead of a constrained problem;
and third, it allows the encoding of our knowledge of what unbiased outputs should be.
Practically, the second allows us to avoid unstable constrained optimisation procedures and to reuse off-the-shelf toolboxes.
The latter translates to the ability to control the level of fairness by
directly varying fairness target rates. 
In contrast, existing approaches rely on intermediate, arguably unintuitive, control parameters such as covariance thresholds.

\end{abstract}



\section{Introduction}
Algorithmic assessment methods are used for predicting human outcomes in areas such as financial services, recruitment, crime and justice, and local government.
This contributes, in theory, to a world with decreasing human biases.
To achieve this, however, we need fair machine learning models that take biased datasets,
but output non-discriminatory decisions to people with differing protected attributes such as gender and marital status.
Datasets can be biased because of, for example, sampling bias, subjective bias of individuals, and institutionalised biases \citep{OltCasDiaKic19,Tol19}.
Uncontrolled bias in the data can translate into bias in machine learning models.

There is no single accepted definition of algorithmic fairness for automated decision-making
but several have been proposed.
One definition is referred to as \emph{statistical} or \emph{demographic parity}.
Given a binary protected attribute (e.g.\ married/unmarried) and a binary decision (e.g.\ yes/no to getting a loan),
demographic parity requires equal positive rates (PR) across the two sensitive groups (married and \emph{un}married individuals should be equally likely to receive a loan).
Another fairness criterion, \emph{equalised odds}~\citep{hardt2016equality},
takes into account the binary decision, and instead of equal PR requires equal true positive rates (TPR) and false positive rates (FPR).
This criterion is intended to be more compatible with the goal of building accurate predictors or achieving high utility~\citep{hardt2016equality}.
We discuss the suitability of the different fairness criteria in the discussion section at the end of the paper.

There are many existing models for enforcing demographic parity and equalised odds
\citep{CreMadJacWeietal19,AgaBeyDudLanetal18,calders2009building,kamishima2012fairness,zafar2017fairnesstreatment,zafar2017fairnessconstraints}.
However, these existing approaches to balancing accuracy and fairness rely on intermediate, unintuitive control parameters such as allowable constraint violation $\epsilon$ (e.g. $0.01$) in \citet{AgaBeyDudLanetal18}, or a covariance threshold $c$
(e.g. $0$ that is controlled by another parameters $\tau$ and $\mu$ -- $0.005$ and $1.2$ -- to trade off this threshold and accuracy)
in \citet{zafar2017fairnesstreatment}.
This is related to the fact that many of these approaches embed fairness criteria as \emph{constraints} in the optimisation procedure
\citep{DonOneBenShaetal18,quadrianto2017recycling,zafar2017fairnesstreatment,zafar2017fairnessconstraints}.

In contrast, we provide a probabilistic classification framework with bias controlling mechanisms that can be tuned based on positive rates (PR),
an intuitive parameter.
Thus, giving humans the control to set the rate of positive predictions (e.g.\ a PR of $0.6$).
Our framework is based on the concept of a \emph{balanced dataset}
and introduces latent target labels, which, instead of the provided labels, are now the training label of our classifier.
We prove bounds on how far the target labels diverge from the dataset labels.
We instantiate our approach with a parametric logistic regression classifier and a Bayesian non-parametric Gaussian process classifier (GPC).
As our formulation is not expressed as a constrained problem,
we can draw upon advancements in automated variational inference~\citep{bonilla2016generic,gardner2018gpytorch,krauth2016autogp}
for learning the fair model, and for handling large amounts of data.

The method presented in this paper is closely related to a number of previous works,
e.g.\ \citet{kamiran2012data,calders2010three}.
Proper comparison with them requires knowledge of our approach.
We will thus explain our approach in the subsequent sections, and defer detailed comparisons to Section~\ref{sec:relatedwork} (Related Work).

\section{Target labels for tuning group fairness}
We will start by describing several notions of group fairness.
For each individual, we have a vector of non-sensitive attributes $x\in\mathcal{X}$, a class label $y\in\mathcal{Y}$, and a sensitive attribute $s\in\mathcal{S}$ (e.g.\ racial origin or gender).
We focus on the case where $s$ and $y$ are binary. 
We assume that a positive label $y=1$ corresponds to a positive outcome for an individual -- for example, being accepted for a loan.
\emph{Group fairness} balances a certain condition between groups of individuals with different sensitive attribute, $s$ versus $s\sp{\prime}$. 
The term $\hat{y}$ below is the prediction of a machine learning model that,
in most works, uses only non-sensitive attributes $x$.
Several group fairness criteria have been proposed (e.g. \citet{zafar2017fairnesstreatment,chouldechova2017fair,hardt2016equality}):
\begin{align}
& \text{equality of positive rate (Demographic Parity):}\nonumber\\
& \text{Pr}(\hat{y}=1|s)=\text{Pr}(\hat{y}=1|s\sp{\prime}) \label{eq:eq_dem_par}\\
& \text{equality of accuracy:}\nonumber\\
& \text{Pr}(\hat{y}=y|s)=\text{Pr}(\hat{y}=y|s\sp{\prime}) \\
& \text{equality of true positive rate (Equality of Opportunity):} \nonumber\\
& \text{Pr}(\hat{y}=1|s,y=1)=\text{Pr}(\hat{y}=1|s\sp{\prime},y=1)  \label{eq:eq_opp_cri}~.
\end{align}
\emph{Equalised odds} criterion corresponds to Equality of Opportunity (\ref{eq:eq_opp_cri}) plus equality of false positive rate.

The Bayes-optimal classifier only satisfies these criteria if the training data itself satisfies them.
That is, in order for the Bayes-optimal classifier to satisfy \emph{demographic parity}, the following must hold:
$\PP(y=1|s) = \PP(y=1|s')$, where $y$ is the training label.
We call a dataset for which $\PP(y, s)=\PP(y)\PP(s)$ holds, a \emph{balanced} dataset.
Given a balanced dataset, a Bayes-optimal classifier learns to satisfy demographic parity
and an approximately Bayes-optimal classifier should learn to satisfy it at least approximately.
Here, we motivated the importance of balanced datasets via the demographic parity criterion,
but it is also important for \emph{equality of opportunity} which we discuss in Section~\ref{ssec:eqopp}.

In general, however, our given dataset is likely to be imbalanced.
There are two common solutions to this problem:
either pre-process or massage the dataset to make it balanced,
or constrain the classifier to give fair predictions despite it having been trained on an unbalanced dataset.
Our approach takes parts from both solutions.

An imbalanced dataset can be turned into a balanced dataset
by either changing the class labels $y$ or the sensitive attributes $s$.
In the use cases that we are interested in,
$s$ is considered an integral part of the input, representing trustworthy information and thus should not be changed.
$y$, conversely, is often not completely trustworthy; it is not an integral part of the sample but merely an observed outcome.
In a hiring dataset, for instance, $y$ might represent the hiring decision, which can be biased,
and not the relevant question of whether someone makes a good employee.

Thus, we introduce new \emph{target labels} $\bar{y}$
such that the dataset is balanced: $\PP(\bar{y}, s)=\PP(\bar{y})\PP(s)$.
The idea is that these target labels still contain as much information as possible about the task,
while also forming a balanced dataset.
This introduces the concept of the accuracy-fairness trade-off:
in order to be completely accurate with respect to the original (not completely trustworthy) class labels $y$, we would require $\bar{y} =y$,
but then, the fairness constraints would not be satisfied.

Let $\eta_s(x)=\PP(y=1|x, s)$ denote the distribution of $y$ in the data.
The target distribution $\bar{\eta}_s(x)=\PP(\bar{y}=1|x, s)$ is then given by
\begin{align}
  \bar{\eta}_s(x)=(\PP(\bar{y}=1|y=1,s) + \PP(\bar{y}=0|y=0,s) - 1)\eta_s(x) + 1 - \PP(\bar{y}=0|y=0,s)%
  \label{eq:etaeta}
\end{align}
due to the marginalisation rules of probabilities.
The conditional probability $\PP(\bar{y}|y,s)$ indicates with which probability we want to keep the class label.
This probability could in principle depend on $x$ which would enable the realisation of individual fairness.
The dependence on $x$ has to be prior knowledge as it cannot be learned from the data.
This prior knowledge can encode the semantics that ``similar individuals should be treated similarly''~\citep{dwork2012fairness},
or that ``less qualified individuals should not be preferentially favoured over more qualified individuals''~\citep{JosKeaMorRot16}.
Existing proposals for guaranteeing individual fairness require strong assumptions,
such as the availability of an agreed-upon similarity metric, or knowledge of the underlying data generating process.
In contrast, in group fairness, we partition individuals into protected groups based on some sensitive attribute $s$
and ask that some statistics of a classifier be approximately equalised across those groups (see (\ref{eq:eq_dem_par})--(\ref{eq:eq_opp_cri})). 
In this case, $\PP(\bar{y}|y,s)$ does not depend on $x$.

Returning to \eqref{eq:etaeta}, we can simplify it with
\begin{align}
  m_s &:= \PP(\bar{y}=1|y=1,s) + \PP(\bar{y}=0|y=0,s) - 1\\
  b_s &:= 1 - \PP(\bar{y}=0|y=0,s)~,
\end{align}
arriving at $\bar{\eta}_s(x)=m_s\cdot \eta_s(x) + b_s$.
$m_s$ and $b_s$ are chosen such that $\PP(\bar{y}, s)=\PP(\bar{y})\PP(s)$.
This can be interpreted as shifting the decision boundary depending on $s$ so that the new distribution is balanced.

As there is some freedom in choosing $m_s$ and $b_s$, it is important to consider what the effect of different values is.
The following theorem provides this (the proof can be found in the Supplementary Material):

\begin{theorem}\label{th:prob}
  The probability that $y$ and $\bar{y}$ disagree ($y\neq\bar{y}$) for any input $x$ in the dataset is given by:
  \begin{align}
    \PP(y\neq\bar{y}|s)=\PP\left(\left|\eta(x,s) - \tfrac{1}{2}\right| < t_s\right)
  \end{align}
  where
  \begin{align}
    t_s = \left|\frac{m_s+2b_s-1}{2m_s}\right|~.\label{eq:def-ts}
  \end{align}
\end{theorem}
Thus, if the threshold $t_s$ is small,
then only if there are inputs very close to the decision boundary ($\eta_s(x)$ close to $\tfrac{1}{2}$)
would we have $\bar{y}\neq y$.
$t_s$ determines the accuracy penalty that we have to accept in order to gain fairness.
The value of $t_s$ can be taken into account when choosing $m_s$ and $b_s$ (see Section~\ref{sec:fairness}).
If $\eta_s$ satisfies the Tsybakov condition~\citep{tsybakov2004optimal},
then we can give an upper bound for the probability.
\begin{definition}
  A distribution $\eta$ satisfies the Tsybakov condition if there exist $C>0$, $\lambda > 0$ and $t_0\in (0,\tfrac{1}{2}]$
  such that for all $t\leq t_0$,
  \begin{align}
    \PP\left(\left|\eta(x)-\tfrac{1}{2}\right|<t\right)\leq Ct^\lambda~.
  \end{align}
\end{definition}
This condition bounds the region close to the decision boundary.
It is a property of the dataset.
\begin{corollary}\label{th:upperbound}
  If $\eta(x,s)=\PP(y=1|x,s)$ satisfies the Tsybakov condition in $x$, with constants $C$ and $\lambda$,
  then the probability that $y$ and $\bar{y}$ disagree ($y\neq\bar{y}$) for any input $x$ in the dataset is bounded by:
  \begin{align}
    \PP(y\neq\bar{y}|s)<C\left|\frac{m_s+2b_s-1}{2m_s}\right|^\lambda~.
  \end{align}
\end{corollary}
Section~\ref{sec:fairness} discusses how to choose the parameters for $\bar{\eta}$ in order to make it balanced.

\subsection{Equality of Opportunity}\label{ssec:eqopp}
In contrast to demographic parity,
equality of opportunity (just as equality of accuracy) is satisfied by a perfect classifier.
Imperfect classifiers, however, do not by default satisfy it:
the true positive rate (TPR) is different for different subgroups.
The reason for this is that while the classifier is optimised to have a high TPR overall,
it is not optimised to have the same TPR in the subgroups.

The overall TPR is a weighted sum of the TPRs in the subgroups:
\begin{align}
  \mathit{TPR}= \PP(s=0|y=1) \cdot \mathit{TPR}_{s=0} + \PP(s=1|y=1) \cdot \mathit{TPR}_{s=1}~.\label{eq:tpr-weighted}
\end{align}
In datasets where the positive label $y=1$ is heavily skewed toward one of the groups
(say, group $s=1$; meaning that $\PP(s=1|y=1)$ is high and $\PP(s=0|y=1)$ is low),
overall TPR might be maximised by setting the decision boundary
such that nearly all samples in $s=0$ are classified as $y=0$,
while for $s=1$ a high TPR is achieved.
The low TPR for $s=0$ is in this case weighted down and only weakly impacts the overall TPR\@.
For $s=0$, the resulting classifier uses $s$ as a shorthand for $y$, mostly ignoring the other features.
This problem usually persists even when $s$ is removed from the input features
because $s$ is implicit in the other features.

A \emph{balanced} dataset helps with this issue
because in such datasets, $s$ is not a useful proxy for the balanced label $\bar{y}$
(because we have $\PP(\bar{y}, s)=\PP(\bar{y})\PP(s)$) and $s$ cannot be used as a shorthand.
Assuming the dataset is balanced in $s$ ($\PP(s=0)=\PP(s=1)$),
for such datasets $\PP(s=0|y=1)=\PP(s=1|y=1)$ holds and the two terms in \eqref{eq:tpr-weighted} have equal weight.

Here as well there is an accuracy-fairness trade-off:
assuming the unconstrained model is as accurate as its model complexity allows,
adding additional constraints like equality of opportunity can only make the accuracy worse.

\subsection{Concrete algorithm}
For training, we are only given the unbalanced distribution $\eta_s(x)$
and not the target distribution $\bar{\eta}_s(x)$.
However, $\bar{\eta}_s(x)$ is needed in order to train a fair classifier.
One approach is to explicitly change the labels $y$ in the dataset, in order to construct $\bar{\eta}_s(x)$.
We discuss this approach and its drawback in the related work section (Section~\ref{sec:relatedwork}).

We present a novel approach which only implicitly constructs the balanced dataset.
This framework can be used with any likelihood-based model, such as Logistic Regression and Gaussian Process models.
The relation presented in \eqref{eq:etaeta} allows us to formulate a likelihood
that targets $\bar{\eta}_s(x)$ while only having access to the imbalanced labels $y$.
As we only have access to $y$, $\PP(y|x,s,\theta)$ is the likelihood to optimise.
It represents the probability that $y$ is the imbalanced label,
given the input $x$, the sensitive attribute $s$ that available in the training set
and the model parameters $\theta$ for a model that is targeting $\bar{y}$.
Thus, we get
\begin{align}
  &\PP(y=1|x, s, \theta)
  = \sum\limits_{\bar{y}\in \{0, 1\}} \PP(y=1,\bar{y}|x, s, \theta)
  =\,\, \sum\limits_{\bar{y}\in \{0, 1\}} \PP(y=1|\bar{y}, x, s, \theta)\,\PP(\bar{y}|x, s, \theta) \label{eq:lik}~.
\end{align}
As we are only considering group fairness, we have $\PP(y=1|\bar{y}, x, s, \theta)=\PP(y=1|\bar{y}, s)$.

Let $f_\theta(x, y')$ be the likelihood function of a given model,
where $f$ gives the likelihood of the label $y'$ given the input $x$ and the model parameters $\theta$.
As we do not want to make use of $s$ at test time, $f$ does not explicitly depend on $s$.
The likelihood with respect to $\bar{y}$ is then given by $f$: $\PP(\bar{y}|x, s, \theta) = f_\theta(x, \bar{y})$;
and thus, does not depend on $s$. 
The latter is important in order to avoid \emph{direct discrimination}~\citep{BarSel16}.
With these simplifications, the expression for the likelihood becomes
\begin{align}
  \PP(y=1|x, s, \theta)
  =\sum\limits_{\bar{y}\in \{0, 1\}} \PP(y=1|\bar{y}, s)\,\PP(\bar{y}|x, \theta) \label{eq:lik2}~.
\end{align}
The conditional probabilities, $\PP(y|\bar{y}, s)$, are closely related to the conditional probabilities in \eqref{eq:etaeta}
and play a similar role of ``transition probabilities''.
Section~\ref{sec:fairness} explains how to choose these transition probabilities in order to arrive at a balanced dataset.
For a binary sensitive attribute $s$ (and binary label $y$), there are 4 transition probabilities
(see Algorithm~\ref{alg:fair} where $d^{s=j}_{\bar{y}=i} := \PP(y=1|\bar{y}=i, s=j)$):
\begin{align}
  &\PP(y=1|\bar{y}=0, s=0),  &&\PP(y=1|\bar{y}=1, s=0) \label{eq:par1}\\
  &\PP(y=1|\bar{y}=0, s=1),  &&\PP(y=1|\bar{y}=1, s=1) \label{eq:par4}~.
\end{align}

A perhaps useful interpretation of \eqref{eq:lik2} is that,
even though we don't have access to $\bar{y}$ directly,
we can still compute the expectation value over the possible values of $\bar{y}$.

The above derivation applies to binary classification but can easily be extended to the multi-class case.

\begin{algorithm}[tb]
  \caption{Fair learning with target labels $\bar{y}$}%
  \label{alg:fair}
  \begin{algorithmic}[1]
    \REQUIRE Training set $\mathcal{D} = \{(x_i, y_i, s_i)\}^N_{i=1}$, transition probabilities $d^{s=0}_{\bar{y}=0}$,
             $d^{s=0}_{\bar{y}=1}$, $d^{s=1}_{\bar{y}=0}$, $d^{s=1}_{\bar{y}=1}$
    \ENSURE fair model parameters $\theta$
    \STATE Initialise $\theta$ (randomly)
    \FORALL{$x_i$, $y_i$, $s_i$}
      \STATE $\PP_{\bar{y}=1} \gets$ $\bar{\eta}(x_i,\theta)$ (e.g. $\text{logistic}(\left\langle x,\theta\right\rangle)$)
      \STATE $\PP_{\bar{y}=0} \gets 1 - \PP_{\bar{y}=1}$
      \IF{$s_i = 0$}
        \STATE $\PP_{y=1} \gets d^{s=0}_{\bar{y}=0} \cdot \PP_{\bar{y}=0} + d^{s=0}_{\bar{y}=1} \cdot \PP_{\bar{y}=1}$
      \ELSE
        \STATE $\PP_{y=1} \gets d^{s=1}_{\bar{y}=0} \cdot \PP_{\bar{y}=0} + d^{s=1}_{\bar{y}=1} \cdot \PP_{\bar{y}=1}$
      \ENDIF
      \STATE $\ell \gets y_i \cdot \PP_{y=1} + (1-y_i) \cdot (1- \PP_{y=1})$
      \STATE update $\theta$ to maximise likelihood $\ell$
    \ENDFOR
  \end{algorithmic}
\end{algorithm}

\section{Transition probabilities for a balanced dataset}\label{sec:fairness}
This section focuses on how to set values of the transition probabilities in order to arrive at balanced datasets.

\subsection{Meaning of the parameters}
Before we consider concrete values, we give some intuition for the transition probabilities.
Let $s=0$ refer to the protected group.
For this group, we want to make more positive predictions than the training labels indicate.
Variable $\bar{y}$ is supposed to be our target proxy label.
Thus, in order to make more positive predictions, some of the $y=0$ labels should be associated with $\bar{y}=1$.
However, we do not know which.
So, if our model predicts $\bar{y} = 1$ (high $\PP(\bar{y}=1|x,\theta)$) while the training label is $y=0$,
then we allow for the possibility that this is actually correct.
That is, $\PP(y=0|\bar{y}=1,s=0)$ is not $0$.
If we choose, for example, $\PP(y=0|\bar{y}=1,s=0)=0.3$
then that means that 30\% of positive target labels $\bar{y} =1$ may correspond to negative training labels $y=0$.
This way we can have more $\bar{y}=1$ than $y=1$, overall.
On the other hand, predicting $\bar{y}=0$ when $y=1$ holds, will always be deemed incorrect:
$\PP(y=1|\bar{y}=0,s=0)=0$;
this is because we do not want any additional negative labels.

For the non-protected group $s=1$, we have the exact opposite situation.
If anything, we have too many positive labels.
So, if our model predicts $\bar{y} = 0$ (high $\PP(\bar{y}=0|x,\theta)$) while the training label is $y=1$,
then we should again allow for the possibility that this is actually correct.
That is, $\PP(y=1|\bar{y}=0,s=1)$ should not be $0$.
On the other hand, $\PP(y=0|\bar{y}=1,s=1)$ should be $0$ because we do not want additional positive labels for $s=1$.
It could also be that the number of positive labels is exactly as it should be,
in which case we can just set $y=\bar{y}$ for all data points with $s=1$.


\subsection{Choice of parameters}\label{sec:dp}
A balanced dataset is characterised by an independence of the label $\bar{y}$ and the sensitive attribute $s$.
Given that we have complete control over the \emph{transition probabilities},
we can ensure this independence by requiring $\PP(\bar{y}=1|s=0)=\PP(\bar{y}=1|s=1)$.
Our constraint is then that both of these probabilities are equal to the same value,
which we will call the target rate $\mathit{PR}_t$ (``PR'' as \emph{positive rate}):
\begin{align}
  &\PP(\bar{y}=1|s=0) \overset{!}{=} \mathit{PR}_t\quad\text{and}
  \quad \PP(\bar{y}=1|s=1) \overset{!}{=} \mathit{PR}_t~.
\end{align}
This leads us to the following constraints for $s'\in\{0, 1\}$:
\begin{align}
  \mathit{PR}_t &= \PP(\bar{y}=1|s=s') 
  =\sum\limits_y \PP(\bar{y}=1|y,s=s')\, \PP(y|s=s').\label{eq:dpconstraint}
\end{align}
We call $\PP(y=1|s=j)$ the base rate $\mathit{PR}_b^j$ which we estimate from the training set:
\begin{align*}
  \PP(y=1|s=i) = \frac{\text{number of points with } y=1 \text{ in group }i}
  {\text{number of points in group }i}~.
\end{align*}
Expanding the sum, we get
\begin{align}
  \mathit{PR}_t =\,\, &\PP(\bar{y}=1|y=0,s=s') \cdot (1-\mathit{PR}_b^1) 
  +\PP(\bar{y}=1|y=1,s=s') \cdot \mathit{PR}_b^1~.\label{eq:dpconstexpanded}
\end{align}
This is a system of linear equations consisting of two equations (one for each value of $s'$)
and four free variables: $\PP(\bar{y}=1|y,s)$ with $y,s\in\{0, 1\}$.
The two unconstrained degrees of freedom determine how strongly the accuracy will be affected by the fairness constraint.
If we set $\PP(\bar{y}=1|y=1,s)$ to 0.5,
then this expresses the fact that a train label $y$ of $1$ only implies a target label $\bar{y}$ of $1$ in 50\% of the cases.
In order to minimise the effect on accuracy,
we make $\PP(\bar{y}=1|y=1,s)$ as high as possible and $\PP(\bar{y}=1|y=0,s)$, conversely, as low as possible.
However, the lowest and highest possible values are not always 0 and 1 respectively.
To see this, we solve for $\PP(\bar{y}=1|y=0,s=j)$ in \eqref{eq:dpconstexpanded}:
\begin{align}
  &\PP(\bar{y}=1|y=0,s=j) 
  =\,\, \frac{\mathit{PR}_b^j}{1-\mathit{PR}_b^j} \left(\frac{\mathit{PR}_t}{\mathit{PR}_b^j} - \PP(\bar{y}=1|y=1,s=j)\right)~.
\end{align}
If $\nicefrac{\mathit{PR}_t}{\mathit{PR}_b^j}$ were greater than 1,
then setting $\PP(\bar{y}=1|y=0,s=j)$ to 0
would imply a $\PP(\bar{y}=1|y=1,s=j)$ value greater than 1.
A visualisation that shows why this happens can be found in the Supplementary Material.
We thus arrive at the following definitions:
\begin{align}
  \PP(\bar{y}=1|y=1,s=j)&=\begin{cases}
    1 &\text{if }\mathit{PR}_t>\mathit{PR}_b^j\\
    \frac{\mathit{PR}_t}{\mathit{PR}_b^j} &\text{otherwise.}
  \end{cases}%
  \label{eq:tau-11}\\
  \PP(\bar{y}=1|y=0,s=j)&=\begin{cases}
    \frac{\mathit{PR}_t-\mathit{PR}_b^j}{1-\mathit{PR}_b^j} &\text{if }\mathit{PR}_t>\mathit{PR}_b^j\\
    0 &\text{otherwise.}
  \end{cases}%
  \label{eq:tau-01}
\end{align}
Algorithm~\ref{alg:parity} shows pseudocode of the procedure, including the computation of the allowed minimal and maximal value.

Once all these probabilities have been found, the transition probabilities needed for \eqref{eq:lik2}
are fully determined by applying Bayes' rule:
\begin{align}
  \PP(y=1|\bar{y}, s) = \frac{\PP(\bar{y}|y=1, s)
  \PP(y=1|s)}{\PP(\bar{y}|s)}~. \label{eq:debias}
\end{align}

\subsubsection{Choosing a target rate.}
\noindent As shown, there is a remaining degree of freedom when targeting a balanced dataset:
the target rate $\mathit{PR}_t := \PP(\bar{y}=1)$.
This is true for both fairness criteria that we are targeting.
The choice of targeting rate affects how much $\eta$ and $\bar{\eta}$ differ as implied by Theorem~\ref{th:prob}
($\mathit{PR}_t$ affects $m_s$ and $b_s$).
$\bar{\eta}$ should remain close to $\eta$
as $\bar{\eta}$ only represents an auxiliary distribution that does not have meaning on its own.
The threshold $t_s$ in Theorem~\ref{th:prob} (\eqref{eq:def-ts}) gives an indication of how close the distributions are.
With the definitions in \eqref{eq:tau-11} and \eqref{eq:tau-01},
we can express $t_s$ in terms of the target rate and the base rate:
\begin{align}
  t_s = \begin{cases}
    \frac{1}{2}\frac{\mathit{PR}_b^s - \mathit{PR}_t}{\mathit{PR}_t} &\text{if }\mathit{PR}_t>\mathit{PR}_b^j\\
    \frac{1}{2}\frac{\mathit{PR}_t - \mathit{PR}_b^s}{1 - \mathit{PR}_t} &\text{otherwise.}
  \end{cases}\label{eq:ts-pr}
\end{align}
This shows that $t_s$ is smallest when $\mathit{PR}_b^s$ and $\mathit{PR}_t$ are closest.
However, as $\mathit{PR}_b^s$ has different values for different $s$,
we cannot set $\mathit{PR}_b^s=\mathit{PR}_t$ for all $s$.
In order to keep both $t_{s=0}$ and $t_{s=1}$ small,
it follows from \eqref{eq:ts-pr} that $\mathit{PR}_t$ should at least be between $\mathit{PR}_b^0$ and $\mathit{PR}_b^1$.
A more precise statement can be made when we explicitly want to minimise the sum $t_{s=0} + t_{s=1}$:
assuming $\mathit{PR}_b^0<\mathit{PR}_t<\mathit{PR}_b^1$ and $\mathit{PR}_b^1<\tfrac{1}{2}$,
the optimal choice for $\mathit{PR}_t$ is $\mathit{PR}_b^1$ (see Supplementary Material for details).
We call this choice $\mathit{PR}_t^{max}$.
For $\mathit{PR}_b^0>\tfrac{1}{2}$, analogous statements can be made,
but this is of less interest as this case does not appear in our experiments.

The previous statements about $t_s$ do not directly translate into observable quantities like accuracy if the Tsybakov condition is not satisfied,
and even if it is satisfied, the usefulness depends on the constants $C$ and $\lambda$.
Conversely, the following theorem makes \emph{generally} applicable statement about the accuracy that can be achieved.
Before we get to the theorem, we introduce some notation.
We are given a dataset $\mathcal{D} = {\{(x_i, y_i)\}}_i$,
where the $x_i$ are vectors of features and the $y_i$ the corresponding labels.
We refer to the tuples $(x, y)$ as the \emph{samples} of the dataset.
The number of samples is $N = |\mathcal{D}|$.

We assume binary labels ($y\in \{0, 1\}$) and thus can form the (disjoint) subsets $\mathcal{\mathcal{Y}}^0$ and $\mathcal{Y}^1$ with
\begin{align}
  \mathcal{Y}^j = \{(x, y)\in \mathcal{D}|y = j\}\quad\text{with } j\in\{0, 1\}~.
\end{align}
Furthermore, we associate each sample with a classification $\hat{y}\in \{0, 1\}$.
The task of making the classification $\hat{y}=0$ or $\hat{y}=1$ can be understood as sorting each sample from $\mathcal{D}$
into one of two sets: $\mathcal{C}^0$ and $\mathcal{C}^1$,
such that $\mathcal{C}^0\cup\mathcal{C}^1 = \mathcal{D}$ and $\mathcal{C}^0\cap\mathcal{C}^1 = \emptyset$.

We refer to the set $\mathcal{A} = (\mathcal{C}^0\cap\mathcal{Y}^0) \cup (\mathcal{C}^1\cap\mathcal{Y}^1)$
as the set of correct (or accurate) predictions.
The \emph{accuracy} is given by $\mathit{acc} = N^{-1}\cdot\left|\mathcal{A}\right|$.
\begin{definition}
  \begin{align}
    r_a := \frac{\left|\mathcal{Y}^1\right|}{|\mathcal{D}|} = \frac{\left|\mathcal{Y}^1\right|}{N}
  \end{align}
  is called the \emph{base acceptance rate} of the dataset $\mathcal{D}$.
\end{definition}

\begin{definition}
  \begin{align}
    \hat{r}_a = \frac{\left|\mathcal{C}^1\right|}{|\mathcal{D}|} = \frac{\left|\mathcal{C}^1\right|}{N}
  \end{align}
  is called the \emph{predictive acceptance rate} of the predictions.
\end{definition}

\begin{theorem}\label{th:fairacc}
  For a dataset with the base rate $r_a$ and corresponding predictions with a predictive acceptance rate of $\hat{r}_a$,
  the accuracy is limited by
  \begin{align}
    \mathit{acc} \leq 1 - \left| \hat{r}_a -r_a \right|~.
  \end{align}
\end{theorem}
\begin{corollary}\label{col:fairacc}
  Given a dataset that consists of two subsets $\mathcal{S}_0$
  and $\mathcal{S}_1$ ($\mathcal{D} = \mathcal{S}_0 \cup \mathcal{S}_1$)
  where $p$ is the ratio of $|\mathcal{S}_0|$ to $|\mathcal{D}|$
  and given corresponding acceptance rates $r^0_a$ and $r^1_a$
  and predictions with target rates $\hat{r}^0_a$ and $\hat{r}^1_a$,
  the accuracy is limited by
  \begin{align}
    \mathit{acc} \leq 1 - p\cdot\left| \hat{r}^0_a -r^0_a \right| - (1-p)\cdot\left| \hat{r}^1_a -r^1_a \right|~.
  \end{align}
\end{corollary}
The proofs are fairly straightforward and can be found in the Supplementary Material.

Corollary~\ref{col:fairacc} implies that
in the common case where group $s=0$ is disadvantaged ($r_a^0 < r_a^1$) and also underrepresented ($p<\tfrac{1}{2}$),
the highest accuracy under demographic parity can be achieved at $\mathit{PR}_t=r_a^1$ with
\begin{align}
  \mathit{acc} \leq 1 - p\cdot\left( r^1_a -r^0_a \right)~.
\end{align}
However, this means willingly accepting a lower accuracy in the (smaller) subset $\mathcal{S}_0$
that is compensated by a very good accuracy in the (larger) subset $\mathcal{S}_1$.
A decidedly ``fairer'' approach is to aim for the same accuracy in both subsets.
This is achieved by using the average of the base acceptance rates for the target rate.
As we balance the test set in our experiments, this kind of sacrificing of one demographic group does not work there.
We compare the two choices ($\mathit{PR}_t^{max}$ and $\mathit{PR}_t^{avg}$) in Section~\ref{sec:experiments}.

\begin{algorithm}[t]
  \caption{Targeting a balanced dataset}%
  \label{alg:parity}
  \begin{algorithmic}[1]
    \REQUIRE target rate $\mathit{PR}_t$,  biased acceptance rate $\mathit{PR}_b^i$
    \ENSURE transition probabilities $d^{s=i}_{\bar{y}=j}$
    \IF{$\mathit{PR}_t > \mathit{PR}_b^i$}
    \STATE $\PP(\bar{y}=1|y=1,s=i) \gets 1$
    \ELSE
    \STATE $\PP(\bar{y}=1|y=1,s=i) \gets \frac{\mathit{PR}_t}{\mathit{PR}_b^i}$
    \ENDIF

    \IF{j=0}
    \STATE $\PP(\bar{y}=0|y=1,s=i) \gets 1 - \PP(\bar{y}=1|y=1,s=i)$
    \STATE $d^{s=i}_{\bar{y}=0} \gets \frac{\PP(\bar{y}=0|y=1,s=i)\cdot \mathit{PR}_b^i}{1 - \mathit{PR}_t}$
    \ELSIF{j=1}
    \STATE $d^{s=i}_{\bar{y}=1} \gets \frac{\PP(\bar{y}=1|y=1,s=i)\cdot \mathit{PR}_b^i}{\mathit{PR}_t}$
    \ENDIF
  \end{algorithmic}
\end{algorithm}

\subsection{Conditionally balanced dataset}
\noindent There is a fairness definition related to demographic parity which allows conditioning on ``legitimate'' risk factors $\ell$
when considering how equal the demographic groups are treated~\cite{corbett2017algorithmic}.
This cleanly translates into balanced datasets which are balanced conditioned on $\ell$:
\begin{align}
  &\PP(\bar{y}=1|\ell=\ell', s=0) \overset{!}{=} \PP(\bar{y}=1|\ell=\ell', s=1)~.
\end{align}
We can interpret this as splitting the data into partitions based on the value of $\ell$,
where the goal is to have all these partitions be balanced.
This can easily be achieved by our method by setting a $\mathit{PR}_t(\ell)$ for each value of $\ell$
and computing the transition probabilities for each sample depending on $\ell$.

\section{Related work}\label{sec:relatedwork}
There are several ways to enforce fairness in machine learning models:
as a pre-processing step
\citep{kamiran2012data,louizos2015variational,lum2016statistical,zemel2013learning,Chiappa19,QuaShaTho19},
as a post-processing step \citep{feldman2015certifying,hardt2016equality}, 
 or as a constraint during the learning phase
\citep{calders2009building,zafar2017fairnesstreatment,zafar2017fairnessconstraints,DonOneBenShaetal18,DimLiuParRad19}.
Our method enforces fairness during the learning phase (an in-processing approach) but, unlike other approaches, we do not cast fair-learning as a \emph{constrained} optimisation problem.
Constrained optimisation requires a customised procedure.
In \citet{GohCotGupFri16}, \citet{zafar2017fairnesstreatment}, and \citet{zafar2017fairnessconstraints},
suitable majorisation-minimisation/convex-concave procedures~\citep{LanSri09} were derived.
Furthermore, such constrained optimisation approaches may lead to more unstable training,
and often yield classifiers with both worse accuracy and more unfair~\citep{CotJiaWanNar18}.

The approaches most closely related to ours were given by \citet{kamiran2012data} who present four pre-processing methods:
\emph{Suppression}, \emph{Massaging the dataset}, \emph{Reweighing}, and \emph{Sampling}.
In our comparison we focus on methods 2, 3 and 4,
because the first one simply removes sensitive attributes and those features that are highly correlated with them.
All the methods given by \citet{kamiran2012data} aim only at enforcing demographic parity.

The massaging approach uses a classifier to first rank all samples according to their probability of having a positive label ($y=1$)
and then flips the labels that are closest to the decision boundary such that the data then satisfies demographic parity.
This \emph{pre-processing} approach is similar in spirit to our \emph{in-processing} method but differs in the execution.
In our method (Section \ref{sec:dp}), ``ranking'' and classification happen in one step and labels are not explicitly flipped
but assigned probabilities of being flipped.

The reweighting method reweights samples based on whether they belong to an over-represented or under-represented demographic group.
The sampling approach is based on the same idea but works by resampling instead of reweighting.
Both reweighting and sampling aim to effectively construct a balanced dataset, without affecting the labels.
This is in contrast to our method which treats the class labels as potentially untrustworthy and allows defying them.

One approach in \citet{calders2010three} is also worth mentioning.
It is based on a \emph{generative} Na\"{i}ve Bayes model in which a latent variable $L$ is introduced
which is reminiscent to our target label $\bar{y}$.
We provide a \emph{discriminative} version of this approach. 
In discriminative models, parameters capture the conditional relationship of an output given an input,
while in generative models, the joint distribution of input-output is parameterised. 
With this conditional relationship formulation
($\PP(y|\bar{y}, s) = \nicefrac{\PP(\bar{y}|y, s) \PP(y|s)}{\PP(\bar{y}|s)} $),
we can have detailed control in setting the target rate.
\citet{calders2010three} focuses only on the demographic parity fairness metric.
\section{Experiments}\label{sec:experiments}
We compare the performance of our target-label model with other existing models based on two real-world datasets.
These datasets have been previously considered in the fairness-aware machine learning literature.

\subsection{Implementation}
The proposed method is compatible with any likelihood-based algorithm. 
%
We consider both a nonparametric and a parametric model.
The nonparametric model is a Gaussian process model, and Logistic regression is the parametric counterpart.
Since our fairness approach is not being framed as a constrained optimisation problem,
we can reuse off-the-shelf toolboxes including the GPyTorch library by \citet{gardner2018gpytorch} for Gaussian process models.
This library incorporates recent advances in scalable variational inference including variational \emph{inducing inputs} and likelihood ratio/REINFORCE estimators.
The variational posterior can be derived from the likelihood and the prior.
We need just need to modify the likelihood to take into account the target labels (Algorithm~\ref{alg:fair}).

\subsection{Data}
We run experiments on two real-world datasets.
The first dataset is the \textbf{Adult Income} dataset~\citep{Dua:2017}.
It contains 33,561 data points with census information from US citizens.
The labels indicate whether the individual earns more ($y=1$) or less ($y=0$) than \$50,000 per year.
We use the dataset with either \emph{race} or \emph{gender} as the sensitive attribute.
The input dimension, excluding the sensitive attributes, is 12 in the raw data;
the categorical features are then one-hot encoded.
For the experiments, we removed 2,399 instances with missing data
and used only the training data, which we split randomly for each trial run.
The second dataset is the \textbf{ProPublica recidivism} dataset.
It contains data from 6,167 individuals that were arrested.
The data was collected when investigating the COMPAS risk assessment tool~\citep{angwin2016machine}.
The task is to predict whether the person was rearrested within two years
($y=1$ if they were rearrested, $y=0$ otherwise).
We again use the dataset with either \emph{race} or \emph{gender} as the sensitive attributes.

\subsection{Balancing the test set}
Any fairness method that is targeting demographic parity, treats the training set as defective in one way:
the acceptance rates are not equal in the training set and this needs to be corrected.
As such, it does not make sense to evaluate these methods on a dataset that is equally defective.
Predicting at equal acceptance rates is the correct result and the test set should reflect this.

In order to generate a test set which has the property of equal acceptance rates,
we subsample the given, imbalanced, test set.
For evaluating demographic parity,
we discard datapoints from the imbalanced test set such that the resulting subset satisfies $\PP(s=j|y=i)=\tfrac{1}{2}$ for all $i$ and $j$.
This balances the set in terms of $s$ and ensures $\PP(y,s)=\PP(y)\PP(s)$,
but does not force the acceptance rate to be $\tfrac{1}{2}$,
which in the case of the Adult dataset would be a severe change as the acceptance rate is naturally quite low there.
Using the described method ensures that the minimal amount of data is discarded for the Adult dataset.
We have empirically observed that all fairness algorithms benefit from this balancing of the test set.

The situation is different for equality of opportunity.
A perfect classifier automatically satisfies equality of opportunity on \emph{any dataset}.
Thus, an algorithm aiming for this fairness constraint should not treat the dataset as defective.
Consequently, for evaluating equality of opportunity we perform no balancing of the test set.

\subsection{Method}
\begin{table*}[tb]
\caption{%
  Accuracy and fairness (with respect to \emph{demographic parity}) for various methods
  on the balanced test set of the Adult dataset.
  Fairness is defined as $\textit{PR}_{s=0}/\textit{PR}_{s=1}$ (a completely fair model would achieve a value of 1.0).
  Left: using \textbf{race} as the sensitive attribute. Right: using \textbf{gender} as the sensitive attribute.
  The mean and std of $10$ repeated experiments.
}%
\label{tab:dempar}
\centering
\resizebox{\textwidth}{!}{
  \begin{tabular}{l@{\hskip 0.25cm}l@{\hskip 0.25cm}l}
    \toprule
  Algorithm &   Fair $\rightarrow 1.0 \leftarrow$&           Accuracy $\uparrow$\\
    \midrule
                   GP &  0.80 $\pm$ 0.07 &  0.888 $\pm$ 0.007 \\
               LR &  0.83 $\pm$ 0.06 &  0.884 $\pm$ 0.007 \\
              SVM &  0.89 $\pm$ 0.06 &  0.899 $\pm$ 0.004 \\
           FairGP (ours) &  0.86 $\pm$ 0.07 &  0.888 $\pm$ 0.006 \\
           FairLR (ours) &  0.90 $\pm$ 0.06 &  0.874 $\pm$ 0.009 \\
    ZafarAccuracy &  0.67 $\pm$ 0.17 &  0.808 $\pm$ 0.016 \\
    ZafarFairness &  0.81 $\pm$ 0.06 &  0.879 $\pm$ 0.009 \\
    \citet{kamiran2012data} &  0.87 $\pm$ 0.07 &  0.882 $\pm$ 0.007 \\
    \citet{AgaBeyDudLanetal18} & 0.86 $\pm$ 0.08 & 0.883 $\pm$ 0.008 \\
    \bottomrule
  \end{tabular}
  \;
  \begin{tabular}{l@{\hskip 0.25cm}l}
    \toprule
    Fair $\rightarrow 1.0 \leftarrow$ &           Accuracy $\uparrow$\\
    \midrule
       0.54 $\pm$ 0.05 &  0.900 $\pm$ 0.006 \\
       0.52 $\pm$ 0.03 &  0.898 $\pm$ 0.003 \\
       0.49 $\pm$ 0.05 &  0.913 $\pm$ 0.004 \\
       0.87 $\pm$ 0.09 &  0.902 $\pm$ 0.007 \\
       0.93 $\pm$ 0.04 &  0.886 $\pm$ 0.012 \\
       0.77 $\pm$ 0.08 &  0.853 $\pm$ 0.017 \\
       0.74 $\pm$ 0.11 &  0.897 $\pm$ 0.004 \\
       0.96 $\pm$ 0.03 &  0.900 $\pm$ 0.004 \\
       0.65 $\pm$ 0.04 &  0.900 $\pm$ 0.004 \\
    \bottomrule
  \end{tabular}
}
\end{table*}%
We evaluate two versions of our target label model\footnote{%
The code 
can be found on GitHub: \url{https://github.com/predictive-analytics-lab/ethicml-models/tree/master/implementations/fairgp}.
}:
\emph{FairGP}, which is based on Gaussian Process models, and \emph{FairLR}, which is based on logistic regression.
We also train baseline models that do not take fairness into account.

In both \emph{FairGP} and \emph{FairLR}, our approach is implemented by modifying the likelihood function.
First, the unmodified likelihood is computed (corresponding to $\PP(\bar{y}=1|x,\theta)$)
and then a linear transformation (dependent on $s$) is applied as given by \eqref{eq:lik2}.
No additional ranking of the samples is needed, because the unmodified likelihood already supplies ranking information.

The fair GP models and the baseline GP model are all based on variational inference and use the same settings.
During training, each batch is equivalent to the whole dataset.
The number of inducing inputs is 500 on the ProPublica dataset
and 2500 on the Adult dataset
which corresponds to approximately $\nicefrac{1}{8}$ of the number of training points for each dataset.
We use a squared-exponential (SE) kernel with automatic relevance determination (ARD)
and the probit function as the likelihood function.
We optimise the hyper-parameters and the variational parameters
using the Adam method~\citep{kingma2014adam} with the default parameters. 
We use the full covariance matrix for the Gaussian variational distribution.
The logistic regression is trained with RAdam~\citep{liu2019variance} and uses L2 regularisation.
For the regularisation coefficient, we conducted a hyper-parameter search over 10 folds of the data.
For each fold, we picked the hyper-parameter which achieved the best fairness among those 5 with
the best accuracy scores.
We then averaged over the 10 hyper-parameter values chosen in this way and then used this average for all runs to obtain our final results.

In addition to the GP and LR baselines, we compare our proposed model with the following methods:
Support Vector Machine (\emph{SVM}), \emph{Kamiran \& Calders}~\citep{kamiran2012data} (``reweighing'' method),
\emph{Agarwal et al.}~\citep{AgaBeyDudLanetal18} (using logistic regression as the classifier)
and several methods given by Zafar et al.~\citep{zafar2017fairnessconstraints,zafar2017fairnesstreatment},
which include maximising accuracy under demographic parity fairness constraints (\emph{ZafarFairness}),
maximising demographic parity fairness under accuracy constraints (\emph{ZafarAccuracy}),
and removing disparate mistreatment by constraining the false negative rate (\emph{ZafarEqOpp}).
Every method is evaluated over 10 repeats that each have different splits of the training and test set.

\subsection{Results for Demographic Parity on Adult dataset}\label{sssec:demparresults}
\begin{figure}[t]
  \begin{center}
    \centerline{\includegraphics[width=0.98\columnwidth]{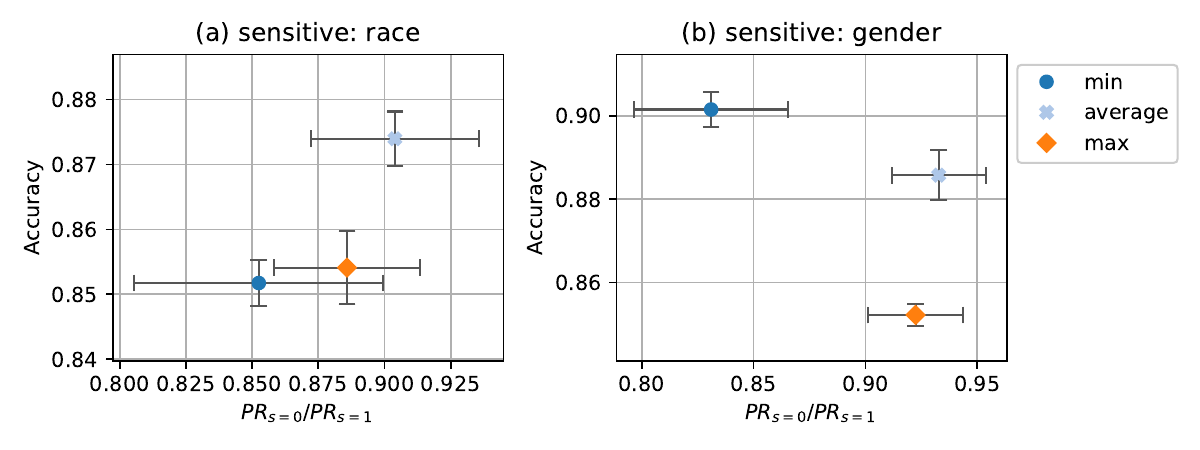}}
    \caption{Accuracy and fairness (demographic parity) for various target choices.
      (a): Adult dataset using race as the sensitive attribute;
      (b): Adult dataset using gender.
      Centre of the cross is the mean; height and width of the box encode half of standard derivation of accuracy and disparate impact.
    }%
    \label{fig:tar}
  \end{center}
\end{figure}%
Following \citet{zafar2017fairnessconstraints} we evaluate demographic parity on the Adult dataset.
Table~\ref{tab:dempar} shows the accuracy and fairness for several algorithms.
In the table, and in the following, we use $\mathit{PR}_{s=i}$ to denote
the observed rate of positive predictions per demographic group $\PP (\hat{y} = 1 | s =i)$.
Thus, $\mathit{PR}_{s=0} / \mathit{PR}_{s=1}$ is a measure for demographic parity,
where a completely fair model would attain a value of $1.0$.
This measure for demographic parity is also called ``disparate impact''
(see e.g.~\citet{feldman2015certifying,zafar2017fairnesstreatment}).
As the results in Table~\ref{tab:dempar} show, FairGP and FairLR are clearly fairer than the baseline GP and LR\@.
We use the mean ($\mathit{PR}_t^{avg}$) for the target acceptance rate.
The difference between fair models and unconstrained models is not as large with \emph{race} as the sensitive attribute,
as the unconstrained models are already quite fair there.
The results of FairGP are characterised by high fairness and high accuracy.
FairLR achieves similar results to FairGP, but with generally slightly lower accuracy but better fairness.
We used the two step procedure of \citet{DonOneBenShaetal18}
to verify that we cannot achieve the same fairness result with just parameter search on LR\@.

In Fig.~\ref{fig:tar}, we investigate which choice of target ($\mathit{PR}_t^{avg}$, $\mathit{PR}_t^{min}$ or $\mathit{PR}_t^{max}$)
gives the best result.
We use $\mathit{PR}_t^{avg}$ for all following experiments as this is the fairest choice (cf. Section~\ref{sec:dp}).
The Fig.\ref{fig:tar}(a) shows results from Adult dataset with \emph{race} as sensitive attribute
where we have $\mathit{PR}_t^{min}=0.156$, $\mathit{PR}_t^{max}=0.267$ and $\mathit{PR}_t^{avg} =0.211$.
$\mathit{PR}_t^{avg}$ performs best in term of the trade-off.
\begin{figure}[t]
  \centering
  \includegraphics[width=0.98\textwidth]{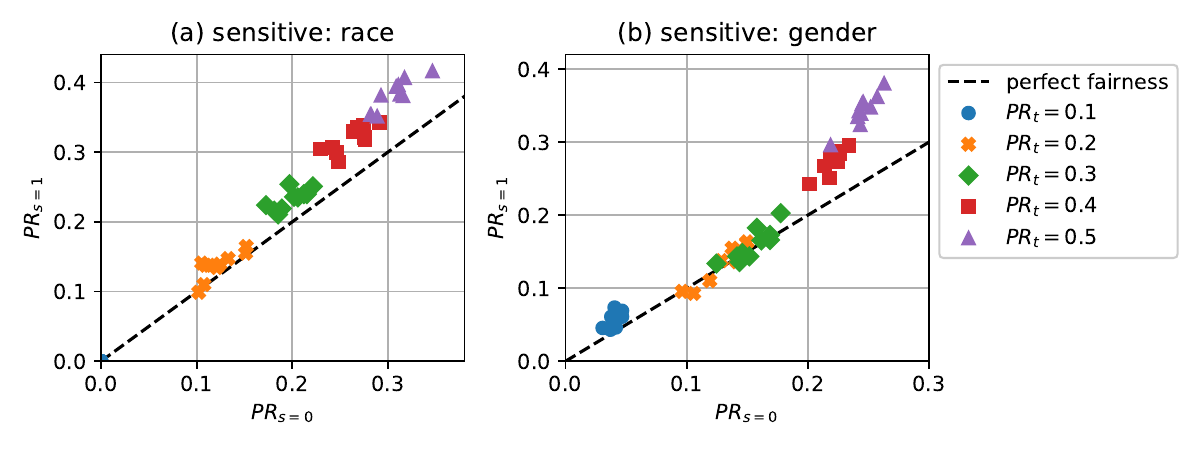}
  \caption{
    Predictions with different target acceptance rates (demographic parity) for 10 repeats.
    (a): $\mathit{PR}_{s=0}$ vs $\mathit{PR}_{s=1}$ using race as the sensitive attribute;
    (b): $\mathit{PR}_{s=0}$ vs $\mathit{PR}_{s=1}$ using gender.
  }%
  \label{fig:adult_parity_scatter_pr_pr}
\end{figure}
\begin{figure}[t]
  \centering
  \includegraphics[width=0.98\textwidth]{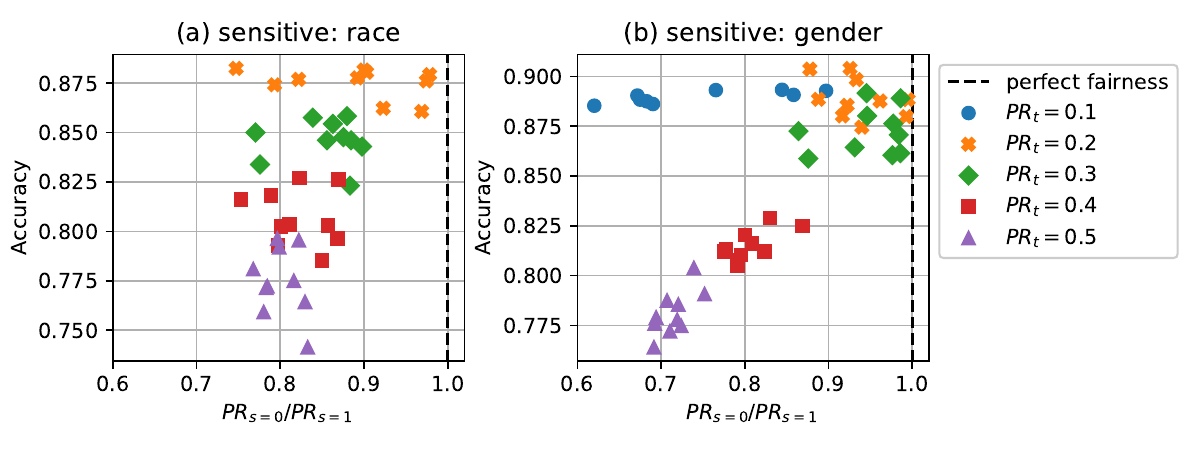}
  \caption{
    Predictions with different target acceptance rates (demographic parity) for 10 repeats.
    (a): disparate impact vs accuracy on Adult dataset using race as the sensitive attribute;
    (b): disparate impact vs accuracy using gender.
  }%
  \label{fig:adult_parity_scatter_acc}
\end{figure}

Fig.~\ref{fig:adult_parity_scatter_pr_pr}(a) and (b) show runs of FairLR
where we explicitly set a target acceptance rate, $\mathit{PR}_t := \PP(\bar{y}=1)$,
instead of taking the mean $\mathit{PR}_t^{avg}$.
A perfect targeting mechanism would produce a diagonal.
The plot shows that setting the target rate has the expected effect on the observed acceptance rate.
This tuning of the target rate is the unique aspect of the approach.
This would be very difficult to achieve with existing fairness methods;
a new constraint would have to be added.
The achieved positive rate is, however, usually a bit lower than the targeted rate (e.g.\ around 0.15 for the target 0.2).
This is due to using imperfect classifiers;
if TPR and TNR differ from 1,
the overall positive rate is affected (see e.g.\ \citet{forman2005counting} for discussion of this).

Fig.~\ref{fig:adult_parity_scatter_acc}(a) and (b) show the same data
as Fig.~\ref{fig:adult_parity_scatter_pr_pr} but with different axes.
It can be seen from this Fig.~\ref{fig:adult_parity_scatter_acc}(a) and (b)
that the fairness-accuracy trade-off is usually best
when the target rate is close to the average of the positive rates in the dataset
(which is around 0.2 for both sensitive attribute).

\subsection{Results for Equality of Opportunity on ProPublica dataset.}\label{sssec:eqoppresults}
%
For equality of opportunity,
we again follow \citet{zafar2017fairnesstreatment} and evaluate the algorithm on the ProPublica dataset.
As we did for demographic parity,
we define a measure of equality of opportunity via the ratio of the true positive rates (TPRs) within the demographic groups.
We use $\mathit{TPR}_{s=i}$ to denote the observed TPR in group $i$: $\PP(\hat{y}=1|y=1, s=i)$,
and $\mathit{TNR}_{s=i}$ for the observed true negative rate (TNR) in the same manner.
The measure is then given by $\mathit{TPR}_{s=0} / \mathit{TPR}_{s=1}$.
A perfectly fair algorithm would achieve $1.0$ on the measure.
\begin{figure*}[t]
  \centering
  \includegraphics[width=0.98\textwidth]{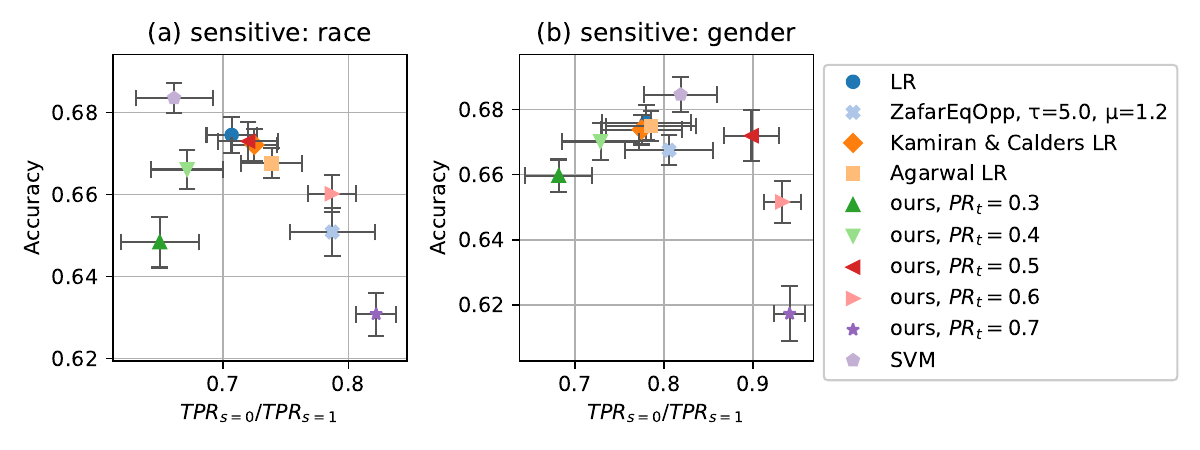}
  \caption{Accuracy and fairness (with respect to \emph{equality of opportunity}) for various methods on ProPublica dataset.
    \textbf{(a)}: using race as the sensitive attribute; \textbf{(b)}: using gender.
    A completely fair model would achieve a value of 1.0 in the x-axis.
    See Fig.~\ref{fig:propublica_opp_scatter_tpr}(a) and (b) on how these choices of PR setting translate to
    $\mathit{TPR}_{s=0}$ vs $\mathit{TPR}_{s=1}$. 
  }%
  \label{fig:propublica_opp_box}
\end{figure*}
\begin{figure*}[!ht]
  \centering
  \includegraphics[width=0.98\textwidth]{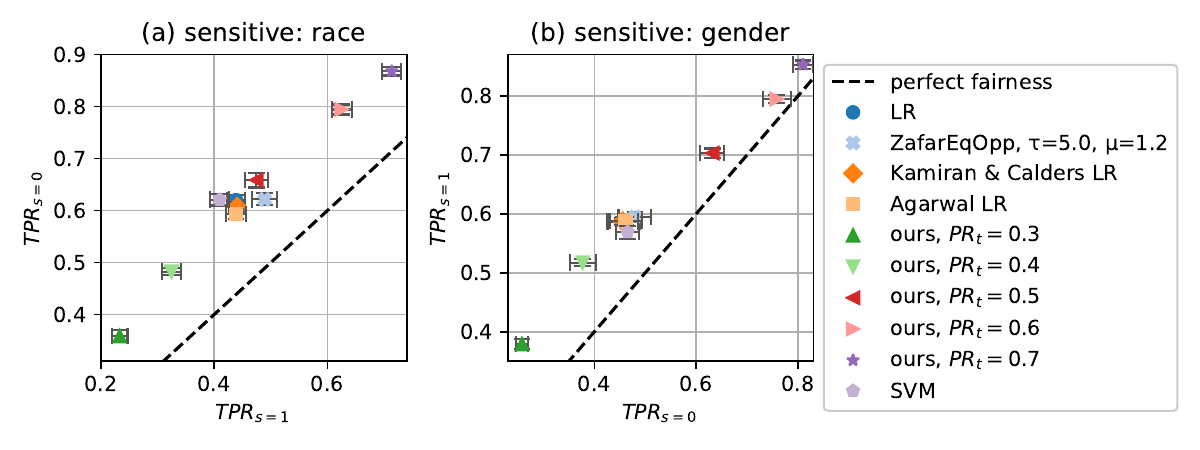}
  \caption{%
    Fairness measure $\mathit{TPR}_{s=0}$ vs $\mathit{TPR}_{s=1}$ (\emph{equality of opportunity}) for different target PRs ($\mathit{PR}_t$).
    (a): on dataset ProPublica recidivism using race as the sensitive attribute;
    (b): using gender.
  }%
  \label{fig:propublica_opp_scatter_tpr}
\end{figure*}

The results of 10 runs are shown in Fig.~\ref{fig:propublica_opp_box} and Fig.~\ref{fig:propublica_opp_scatter_tpr}.
Fig.~\ref{fig:propublica_opp_box}(a) and (b) show the accuracy-fairness trade-off;
Fig.~\ref{fig:propublica_opp_scatter_tpr}(a) and (b) show the achieved TPRs.
In the accuracy-fairness plot, varying $\mathit{PR}_t$ is shown to produce an inverted U-shape:
Higher $\mathit{PR}_t$ still leads to improved fairness, but at a high cost in terms of accuracy.

The latter two plots make clear that the TPR ratio does not tell the whole story:
the realisation of the fairness constraint can differ substantially.
By setting different target PRs for our method, we can affect TPRs as well,
where higher $\mathit{PR}_t$ leads to higher TPR,
stemming from the fact that making more positive predictions increases the chance of making correct positive predictions.
Fig.~\ref{fig:propublica_opp_scatter_tpr} shows that our method can span a wide range of possible TPR values.
Tuning these hidden aspects of fairness is the strength of our method.

\section{Discussion and conclusion}
Fairness is fundamentally not a challenge of algorithms alone, but very much a sociological challenge.
A lot of proposals have emerged recently for defining and obtaining fairness in machine learning-based decision making systems. 
%
The vast majority of academic work has focused on two categories of definitions: statistical (group) notions of fairness and individual notions of fairness (see~\cite{verma2018fairness} for at least twenty different notions of fairness).
Statistical notions are easy to verify but do not provide protections to individuals. 
Individual notions do give individual protections but need strong assumptions, such as the availability of an agreed-upon similarity metric, which can be difficult in practice.
We acknowledge that a proper solution to algorithmic fairness cannot rely on statistics alone.
Nevertheless, these statistical fairness definitions can be helpful in understanding the problem and working towards solutions.
To facilitate this, at every step, the trade-offs that are present should be made very clear
and long-term effects have to be considered as well~\citep{liu2018delayed,kallus2018residual}.

Here, we have developed a machine learning framework which allows us to learn from an implicit balanced dataset,
thus satisfying the two most popular notions of fairness \citep{verma2018fairness}, demographic parity
(also known as \emph{avoiding disparate treatment}) and equality of opportunity (or \emph{avoiding disparate mistreatment}).
Additionally, we indicate how to extend the framework to cover conditional demographic parity as well.
The framework allows us to set a \emph{target rate} to control how the fairness constraint is realised.
For example, we can set the target positive rate for demographic parity to be $0.6$ for different groups.
Depending on the application, it can be important to specify
whether non-discrimination ought to be achieved by more positive predictions or more negative predictions.
This capability is unique to our approach and can be used as an intuitive mechanism to control the realisation of fairness.
Our framework is general and will be applicable for sensitive variables with binary and multi-level values.
The current work focuses on a single binary sensitive variable. 
Future work could extend our tuning approach to other fairness concepts like the closely related predictive parity group fairness~\citep{chouldechova2017fair}
or individual fairness~\citep{dwork2012fairness}.

\section*{Acknowledgements}
Supported by the UK EPSRC project EP/P03442X/1 `EthicalML: Injecting Ethical and Legal Constraints into Machine Learning Models'
and the Russian Academic Excellence Project `5--100'.
We gratefully acknowledge NVIDIA for GPU donations, and Amazon for AWS Cloud Credits.
We thank Chao Chen and Songzhu Zheng for their inspiration of our main proof.



\bibliography{references.bib}

\clearpage
\appendix
\section{Proof of Theorem 1}
Let $\eta(x,s)=P(y=1|x,s)$ be the distribution of the training data.
Let $\bar{\eta}(x, s)=m_s\cdot\eta(x,s)+b_s$, where
\begin{align}
  m_s&=P(\bar{y}=1|y=1,s)-P(\bar{y}=1|y=0,s)\nonumber\\
  &=1-P(\bar{y}=0|y=1,s)-P(\bar{y}=1|y=0,s)\\
  b_s&=P(\bar{y}=1|y=0,s)
\end{align}
So, $\bar{\eta}(x, s)=P(\bar{y}=1|x, s)$.
Let $y$ denote the \emph{hard} labels for $\eta$: $y=\mathbb{I}\left[\eta>\tfrac{1}{2}\right]$ and $\bar{y}$ be the hard labels for $\bar{\eta}$: $\bar{y}=\mathbb{I}\left[\bar{\eta}>\tfrac{1}{2}\right]$.

\begin{theorem}\label{th:tsybakov}
  The probability that $y$ and $\bar{y}$ disagree ($y\neq\bar{y}$) for any input $x$ in the dataset is given by:
  \begin{align}
    \PP(y\neq\bar{y}|s)=\PP\left(\left|\eta(x,s) - \tfrac{1}{2}\right| < t_s\right)
  \end{align}
  where
  \begin{align}
    t_s = \left|\frac{m_s+2b_s-1}{2m_s}\right|~.\label{eq:def-ts2}
  \end{align}
\end{theorem}
\begin{proof}
  The decision boundary that lets us recover the true labels is at $\tfrac{1}{2}$ (independent of $s$).
  So, for the shifted distribution, $\bar{\eta}$, this threshold to get the true labels would be at $\tfrac{1}{2}\cdot m_s + b_s$ (it depends on $s$ now).
  If we however use the decision boundary of $\tfrac{1}{2}$ for $\bar{\eta}$, to make our predictions, $\bar{y}$, then this prediction will sometimes not correspond to the true label, $y\neq\bar{y}$. When does this happen?

  Let $d_s$ be the new decision boundary: $d_s=\tfrac{1}{2}\cdot m_s + b_s$. There are two possibilities to consider here:
  either $\tfrac{1}{2}<d_s$ or $\tfrac{1}{2}>d_s$ (for $d_s=\tfrac{1}{2}$, the decision boundaries are the same and nothing has to be shown).
  The problem, $y\neq\bar{y}$, appears then exactly when the value of $\bar{\eta}$ is between the two boundaries:
  \begin{align}
    \text{if}~d_s>\tfrac{1}{2}\text{:}\quad d_s>\bar{\eta}(x, s)>\tfrac{1}{2} \\
    \text{if}~d_s<\tfrac{1}{2}\text{:}\quad d_s<\bar{\eta}(x, s)<\tfrac{1}{2}
  \end{align}
  Expressing this in terms of $\eta$ and simplifying leads to
  (if $m_s$ is negative, then the two cases are swapped, but we still get both inequalities):
  \begin{align}
    \text{if}~d_s>\tfrac{1}{2}\text{:}\quad\frac{1}{2}>\eta(x,s)>\frac{1-2b_s}{2m_s} \\
    \text{if}~d_s<\tfrac{1}{2}\text{:}\quad\frac{1}{2}<\eta(x,s)<\frac{1-2b_s}{2m_s}
  \end{align}
  This can be summarized as
  \begin{align}
    \left|\eta(x,s)-\frac{1}{2}\right|<\left|\frac{1}{2}-\frac{1-2b_s}{2m_s}\right|~.
  \end{align}
  Let $t_s$ denote the term on the right side of this inequality (i.e. the ``threshold'' that determines whether $y=\bar{y}$ or not). Then
  \begin{align}
    t_s=\left|\frac{1}{2} -\frac{1-2b_s}{2m_s}\right|=\left|\frac{m_s+2b_s-1}{2m_s}\right|~.
  \end{align}
  So, we have: $\left|\eta(x, s) - \frac{1}{2}\right| < t_s =\left|\tfrac{m_s+2b_s-1}{2m_s}\right|$.
  This leads directly to the statement we wanted to prove:
  \begin{align}
    P(y\neq\bar{y}|s)=P\left(\left|\eta(x, s)-\frac{1}{2}\right| <t_s\right)~.
  \end{align}
\end{proof}

\section{Finding minimal $t_s$}
We express $t_s$ in terms of $\mathit{PR}_b^s$ and $\mathit{PR}_t$.
\begin{align}
  t_s = \begin{cases}
    \frac{1}{2}\frac{\mathit{PR}_b^s - \mathit{PR}_t}{\mathit{PR}_t} &\text{if }\mathit{PR}_t>\mathit{PR}_b^j\\
    \frac{1}{2}\frac{\mathit{PR}_t - \mathit{PR}_b^s}{1 - \mathit{PR}_t} &\text{otherwise.}
  \end{cases}\label{eq:ts-pr2}
\end{align}
Without loss of generality, we assume $\mathit{PR}_b^0<\mathit{PR}_b^1$.
As mentioned in the main text, 
$\mathit{PR}_t$ should be between $\mathit{PR}_b^0$ and $\mathit{PR}_b^1$ to minimize both $t_s$.
If that is the case, then we get
\begin{align}
  t_{s=0} &= \frac{1}{2}\frac{\mathit{PR}_t - \mathit{PR}_b^0}{1 - \mathit{PR}_t}\\
  t_{s=1} &= \frac{1}{2}\frac{\mathit{PR}_b^1 - \mathit{PR}_t}{\mathit{PR}_t}~.
\end{align}
If we further assume $\mathit{PR}_b^1<\tfrac{1}{2}$,
then we also have $\mathit{PR}_t<\tfrac{1}{2}$ and thus $\mathit{PR}_t<1-\mathit{PR}_t$.
This implies that the denominator of $t_{s=1}$ is smaller and that, in turn, $t_{s=1}$ grows faster.
This faster growth means that when minimizing $t_{s=0} + t_{s=1}$, we have to concentrate on $t_{s=1}$.
The minimum is then such that $t_{s=1}$ is 0, i.e.\ $\mathit{PR}_t=\mathit{PR}_b^1$.

\section{Proof of Theorem 2}
We are given a dataset $\mathcal{D} = {\{(x_i, y_i)\}}_i$,
where the $x_i$ are vectors of features and the $y_i$ the corresponding labels.
We refer to the tuples $(x, y)$ as the \emph{samples} of the dataset.
The number of samples is $N = |\mathcal{D}|$.

We assume binary labels ($y\in \{0, 1\}$) and thus can form the (disjoint) subsets $\mathcal{\mathcal{Y}}^0$ and $\mathcal{Y}^1$ with
\begin{align}
  \mathcal{Y}^j = \{(x, y)\in \mathcal{D}|y = j\}\quad\text{with } j\in\{0, 1\}~.
\end{align}
Furthermore, we associate each sample with a classification $\hat{y}\in \{0, 1\}$.
The task of making the classification $\hat{y}=0$ or $\hat{y}=1$ can be understood as putting each sample from $\mathcal{D}$
into one of two sets: $\mathcal{C}^0$ and $\mathcal{C}^1$,
such that $\mathcal{C}^0\cup\mathcal{C}^1 = \mathcal{D}$ and $\mathcal{C}^0\cap\mathcal{C}^1 = \emptyset$.

We refer to the set $\mathcal{A} = (\mathcal{C}^0\cap\mathcal{Y}^0) \cup (\mathcal{C}^1\cap\mathcal{Y}^1)$
as the set of correct (or \textbf{a}ccurate) predictions.
The \emph{accuracy} is given by $\mathit{acc} = N^{-1}\cdot\left|\mathcal{A}\right|$.
From the definition it is clear that $0\leq\mathit{acc} \leq 1$.

\begin{definition}
  \begin{align}
    r_a := \frac{\left|\mathcal{Y}^1\right|}{|\mathcal{D}|} = \frac{\left|\mathcal{Y}^1\right|}{N}
  \end{align}
  is called the \emph{acceptance rate} of the dataset $\mathcal{D}$.
\end{definition}

\begin{definition}
  \begin{align}
    \hat{r}_a = \frac{\left|\mathcal{C}^1\right|}{|\mathcal{D}|} = \frac{\left|\mathcal{C}^1\right|}{N}
  \end{align}
  is called the \emph{target rate} of the predictions.
\end{definition}

\begin{theorem}
  For a dataset with the acceptance rate $r_a$ and corresponding predictions with a target rate of $\hat{r}_a$,
  the accuracy is limited by
  \begin{align}
    \mathit{acc} \leq 1 - \left| \hat{r}_a -r_a \right|~.
  \end{align}
\end{theorem}
\begin{proof}
  We first note that by multiplying by $N$, the inequality becomes
  \begin{align}
    |\mathcal{A}| \leq N - \left| |\mathcal{C}^1| -|\mathcal{Y}^1| \right|~.
  \end{align}
  We will choose the predictions $\hat{y}$ that achieve the highest possible accuracy (largest possible $\mathcal{A}$)
  and show that this can never exceed $1 - \left| \hat{r}_a -r_a \right|$.
  As the set $\mathcal{Y}^1$ contains all samples that correspond to $y=1$,
  we try to take as many samples from $\mathcal{Y}^1$ for $\mathcal{C}^1$ as possible.
  Likewise, we take as many indices as possible from $\mathcal{Y}^0$ for $\mathcal{C}^0$.

  We consider three cases: $\hat{r}_a = r_a$, $\hat{r}_a < r_a$ and $\hat{r}_a > r_a$.
  The first case is trivial;
  we have $\left|\mathcal{C}^1\right| = \left|\mathcal{Y}^1\right|$ and thus
  are able to set $\mathcal{C}^1 = \mathcal{Y}^1$, $\mathcal{C}^0 = \mathcal{Y}^0$ and achieve perfect accuracy ($\mathit{acc} \leq 1$).

  For $\hat{r}_a < r_a$, we have $\left|\mathcal{C}^1\right| < \left|\mathcal{Y}^1\right|$
  and thus have more samples available with $y = 1$ than we would optimally need to select for $\mathcal{C}^1$.
  There are two terms to consider that make up the definition of $\mathcal{A}$:
  $\mathcal{C}^0 \cap \mathcal{Y}^0$ and $\mathcal{C}^1\cap \mathcal{Y}^1$.
  The intersection of these two terms is empty because $\mathcal{C}^0\cap \mathcal{C}^1 = \emptyset$.
  Thus,
  \begin{align}
    \left|\mathcal{A}\right| &= \left|(\mathcal{C}^0\cap\mathcal{Y}^0) \cup (\mathcal{C}^1\cap\mathcal{Y}^1)\right|
    = \left|(\mathcal{C}^0\cap\mathcal{Y}^0)\right| + \left|(\mathcal{C}^1\cap\mathcal{Y}^1)\right|~.
  \end{align}
  Selecting samples from $\mathcal{Y}^1$ for $\mathcal{C}^0$ will only \emph{decrease} the first term,
  so for maximum accuracy, it is fine to take as many samples from $\mathcal{Y}^1$ for $\mathcal{C}^1$.
  Taking all available samples from $\mathcal{Y}^1$ such that $\mathcal{C}^1 \supset \mathcal{Y}^1$,
  there is still space left in $\mathcal{C}^1$ which we will have to fill with samples with $y=0$.
  Thus, we have $\mathcal{C}^1\cap \mathcal{Y}^1 = \mathcal{Y}^1$.
  For $\mathcal{C}^0$, we have enough $y=0$ such that
  $\mathcal{C}^0\subset \mathcal{Y}^0$ and $\mathcal{C}^0\cap \mathcal{Y}^0 = \mathcal{C}^0$.
  This is the largest we can make these intersections.
  Putting everything together:
  \begin{align}
    \left|\mathcal{A}^\mathit{optimal}\right|
    &= \left|(\mathcal{C}^0\cap\mathcal{Y}^0)\right| + \left|(\mathcal{C}^1\cap\mathcal{Y}^1)\right|
     = \left|\mathcal{C}^0\right| + \left|\mathcal{Y}^1\right|\nonumber\\
    &= N - \left|\mathcal{C}^1\right| + \left|\mathcal{Y}^1\right|
     = N - \left(\left|\mathcal{C}^1\right| - \left|\mathcal{Y}^1\right|\right)~.
  \end{align}

  For $\hat{r}_a > r_a$, the roles of $\mathcal{C}^0$ and $\mathcal{C}^1$ are reversed
  and thus, the signs in the equation are inverted:
  \begin{align}
    \left|\mathcal{A}^\mathit{optimal}\right| = N  - (\left|\mathcal{Y}^1\right| - \left|\mathcal{C}^1\right|)~.
  \end{align}
  This proves the claim.
\end{proof}
\begin{corollary}
  Given a dataset that consists of two subsets $\mathcal{S}_0$
  and $\mathcal{S}_1$ ($\mathcal{D} = \mathcal{S}_0 \cup \mathcal{S}_1$)
  where $p$ is the ratio of $|\mathcal{S}_0|$ to $|\mathcal{D}|$
  and given corresponding acceptance rates $r^0_a$ and $r^1_a$
  and predictions with target rates $\hat{r}^0_a$ and $\hat{r}^1_a$,
  the accuracy is limited by
  \begin{align}
    \mathit{acc} \leq 1 - p\cdot\left| \hat{r}^0_a -r^0_a \right| - (1-p)\cdot\left| \hat{r}^1_a -r^1_a \right|~.
  \end{align}
\end{corollary}
\begin{figure}[t]
  \centering
  \includegraphics[width=0.5\textwidth]{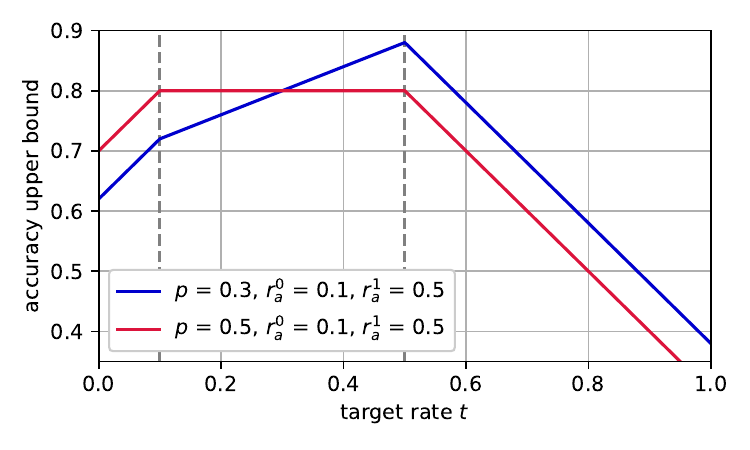}
  \caption{Achievable accuracy for different target values.}%
  \label{fig:accvstarget}
\end{figure}
\begin{example}
  We consider the case where $S_0$
  (which could for example be all data points for female individuals) makes up 30\% of the dataset; so $p = 0.3$.
  Further, we say that for $S_0$ we have an acceptance rate of 10\% ($r^0_a = 0.1$)
  and for $S_1$, 50\% ($r^1_a = 0.5$).
  If we then set both target rates to the same value $t$ ($\hat{r}^0_a=\hat{r}^1_a=t$), with $t = 0.3$,
  then the highest accuracy that can be achieved is $0.8$ or 80\%.

  Fig~\ref{fig:accvstarget} shows the achievable accuracy for different values of $t$ in blue:
  We can see that we can achieve the highest accuracy for $t=r^1_a=0.5$, namely 88\%.
  The plot in orange shows the achievable accuracy for $p=0.5$, i.e., when the two subsets have the same size.
  In this case, all target rates between $r^0_a$ and $r^1_a$ give equal results, namely 80\%.
\end{example}

\section{Illustration of restrictions on PR}
\begin{figure}[t]
  \centering
  \includegraphics[width=0.3\textwidth]{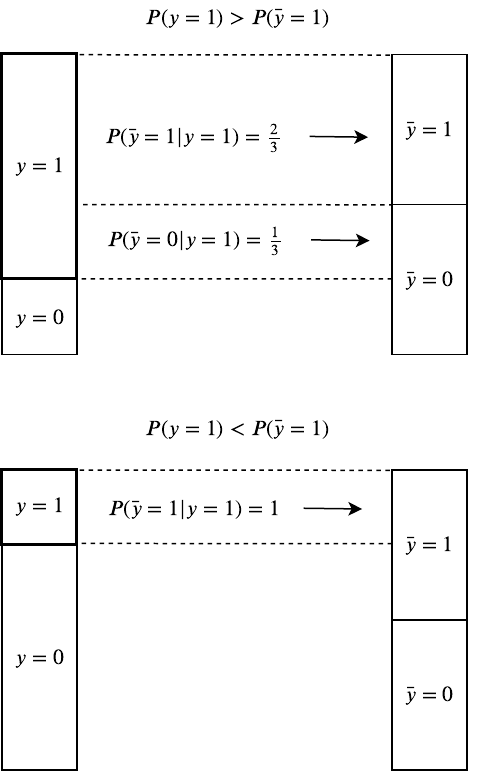}
  \caption{%
    Illustration of demographic parity with target labels.
    In the situation in the upper part, $P(\bar{y}=1|y=1)$ cannot be set to 1,
    because there are more samples with $y=1$ than there are $\bar{y}=1$.
    In the situation in the lower part, $P(\bar{y}=1|y=1)$ can be set to 1.
  }%
  \label{fig:shift}
\end{figure}%
We start by setting a target rate $r_t$:
\begin{align}
  P(\bar{y}=1|s=0) \overset{!}{=} r_t \quad\text{and}\quad P(\bar{y}=1|s=1) \overset{!}{=} r_t
\end{align}

This leads us to the following constraint for $s'\in\{0, 1\}$:
\begin{align}
  r_t &= P(\bar{y}=1|s=s')\nonumber\\
  &= \sum\limits_y P(\bar{y}=1|y,s=s') P(y|s=s')
\end{align}

For $P(y|s=s')$ we will put in the value at which we want our constraint to hold.
We use $r_b^j$ to denote the base rate $P(y=1|s=j)$ which we estimate from the training set.
Plugging this in, we are left with
\begin{align}
  r_t = \,\,&P(\bar{y}=1|y=0,s=0) \cdot (1-r_b^0) \nonumber\\
  +\, &P(\bar{y}=1|y=1,s=0) \cdot r_b^0 \\
  r_t = \,\,&P(\bar{y}=1|y=0,s=1) \cdot (1-r_b^1)\nonumber\\
  +\, &P(\bar{y}=1|y=1,s=1) \cdot r_b^1~.
\end{align}
This is a system of linear equations with two equations and four free variables.
There is thus still considerable freedom in how we want our constraint to be realized.
The freedom that we have here concerns how strongly the accuracy will be affected.

If we set $P(\bar{y}=1|y=1,s)$ to 0.5,
then we express the fact that a train label of 1 only implies a target label of 1 in 50\% of the cases.
In order to minimize the effect on accuracy,
we make $P(\bar{y}=1|y=1,s)$ as high as possible and $P(\bar{y}=1|y=0,s)$ as low as possible.

We solve for $P(\bar{y}=1|y=0,s=j)$:
\begin{align}
  &P(\bar{y}=1|y=0,s=j)\nonumber\\
  =\,\, &\frac{r_b^j}{1-r_b^j} \left(\frac{r_t}{r_b^j} - P(\bar{y}=1|y=1,s=j)\right)~.
\end{align}
However, we can set $P(\bar{y}=1|y=0,s=j)$ to 0
only if that does not imply $P(\bar{y}=1|y=1,s=j)$ will be greater than 1.
This would happen if $\nicefrac{r_t}{r_b^j}$ were greater than 1.

Figure~\ref{fig:shift} illustrates this.
In the upper part of the figure, we have $\nicefrac{r_t}{r_b^j}$ less than 1.
This means, the target positive rate is less than the base positive rate,
and implies that the positive rate has to be lowered somehow.
This is accomplished by mapping some of the $y=1$ samples to $\bar{y}=0$.
Thus, $P(\bar{y}=1|y=1,s=j)$ is less than 1, and $P(\bar{y}=0|y=1,s=j)$ greater than 0.
In the lower part of the figure, we have the opposite case; essentially, $\bar{y}$ and $y$ swap places.
Here, $P(\bar{y}=1|y=1,s=j)=1$ is possible.





\end{document}